%% file: templateArxiv.tex
\documentclass{article}

\usepackage{PRIMEarxiv}

\usepackage[utf8]{inputenc} 
\usepackage[T1]{fontenc}    
\usepackage{hyperref}       
\usepackage{url}            
\usepackage{booktabs}       
\usepackage{amsfonts}       
\usepackage{nicefrac}       
\usepackage{microtype}      
\usepackage{lipsum}
\usepackage{fancyhdr}       
\usepackage{graphicx}       
\usepackage{xcolor}
\usepackage{amsmath}
\usepackage{color}
\usepackage{listings}
\graphicspath{{media/}}     

\title{NerfAcc: A General NeRF Acceleration Toolbox.
}

\author{
  Ruilong Li \\
  University of California, Berkeley \\
  \texttt{ruilongli@berkeley.edu} \\
  \And
  Matthew Tancik \\
  University of California, Berkeley \\
  \texttt{tancik@berkeley.edu} \\
  \And
  Angjoo Kanazawa \\
  University of California, Berkeley \\
  \texttt{kanazawa@berkeley.edu}
}

\begin{document}
\maketitle

\begin{abstract}
We propose NerfAcc, a toolbox for efficient volumetric rendering of radiance fields. We build on the techniques proposed in Instant-NGP~\cite{muller2022instant}, and extend these techniques to not only support bounded static scenes, but also for dynamic scenes and unbounded scenes. NerfAcc comes with a user-friendly Python API, and is ready for plug-and-play acceleration of most NeRFs. Various examples are provided to show how to use this toolbox. Code can be found here: \url{https://github.com/KAIR-BAIR/nerfacc}.\footnote{This write-up matches with NerfAcc v$0.3.5$. For the latest features in NerfAcc, Please check out our more recent write-up at \url{https://arxiv.org/pdf/2305.04966.pdf}}

\end{abstract}


\section{Introduction}
Neural Radiance Fields (NeRFs)~\cite{mildenhall2021nerf} are a groundbreaking technique for 3D representation, 
which model the geometry and view-dependent appearance of the scene by using a multi-layer perceptron (MLP). 
In the past two years, they have been proven to be quite powerful in many downstream applications in 3D such as static/dynamic scene reconstruction~\cite{wang2021neus,yariv2021volume,li2022tava,peng2021animatable}, relighting~\cite{kuang2021neroic,bi2020neural,zhang2021nerfactor,srinivasan2021nerv,zhang2021physg} and content generation~\cite{jain2022zero,poole2022dreamfusion,meng2021gnerf}. 
Using differentiable volumetric rendering~\cite{max1995optical}, a radiance field can be optimized / trained from calibrated RGB images alone. 
However, with the vanilla MLP NeRF, such an optimization process usually takes days to converge on a single scene, and tens of seconds to render a single image.
%

Recently, many works have demonstrated that the NeRF optimization and inference process can be drastically accelerated by incorporating explicit 3D voxels into the radiance field representation~\cite{yu2021plenoctrees,yu2021plenoxels,muller2022instant,chen2022tensorf}. 
%
%
%
%
However, voxel-based radiance fields are less flexible than the MLP-based representation as they are only applicable to a \emph{single}, \emph{static} scene. Applications to
dynamic scene reconstruction~\cite{li2022tava,pumarola2021d,park2021nerfies} and generalization across multiple scenes~\cite{yu2021pixelnerf,wang2021ibrnet,kuang2021neroic} have not benefited yet from those methods.

Despite all variations of the radiance field representations and applications, most of them share the same volumetric rendering process -- cast a ray into the space and accumulate the colors along the ray. 
However, not too much attention has been spent on exploring an efficient volumetric rendering process that is easily applicable to any downstream applications.
A few works such as Plenoctrees~\cite{yu2021plenoctrees} and Instant-NGP~\cite{muller2022instant} incorporate some techniques for efficient volumetric rendering such as skipping empty and occluded areas, but are highly customized for \emph{bounded static} scenes and use specialized CUDA implementations making it difficult to extend to other applications.

In this work, we propose a toolbox, NerfAcc for ``NeRF Accelration'', which focuses on efficient volumetric rendering of radiance fields. We build on the techniques proposed in Instant-NGP~\cite{muller2022instant}, and extend those techniques to support not only bounded static scenes, but also the dynamic scenes and unbounded scenes. We also provide a user friendly Python API written with PyTorch~\cite{paszke2019pytorch}, which is ready for plug-and-play acceleration of most NeRFs. Various examples are included to show how to use the NerfAcc toolbox to accelerate existing NeRF methods.

With the NerfAcc toobox,
\begin{itemize}
    \item The vanilla NeRF model~\cite{mildenhall2021nerf} with an 8-layer MLP can be trained to better quality (+0.5 PSNR) in 1 hour rather than $1 \sim 2$ days as in the paper.
    \item The Instant-NGP NeRF~\cite{muller2022instant} model can be trained to equal quality with 4.5 minutes in Python, comparing to the official pure CUDA implementation.
    \item The D-NeRF~\cite{pumarola2021d} model for dynamic scenes can be trained on 1 hour rather than 2 days as in the paper, and with better quality (+2.0 PSNR).
\end{itemize}

\section{Design}

In NerfAcc, the volumetric rendering pipeline is broken down into two steps, \emph{ray marching} and \emph{differentiable rendering}. Here we dive into each step to introduce some techniques that we incorporated for efficient volumetric rendering.

\subsection{Ray Marching}

Ray marching is the process of casting a ray through the scene and generating discrete samples along the ray. As those samples will later be evaluated by a radiance field, which is usually the main bottleneck of the entire pipeline, efficiency can be achieved by reducing the number of samples as much as possible during ray marching. As shown in the volumetric rendering Equation~\ref{equ:nerf}, a sample would have little contribution to the final image color $C(\mathbf{r})$, if it has either low opacity $\alpha_i$ or low transmittance $T_i$. In other words, during ray marching we can safely skip samples that are in the empty or occluded regions as they do not contribute to the final image.

\begin{align}
\label{equ:nerf}
    C(\mathbf{r}) = \sum_{i=1}^{N} T_i \alpha_i\mathbf{c}_i, 
    \quad \text{where } T_i = \text{exp}(-\sum_{j=1}^{i-1}\sigma_j\delta_j),~ \alpha_i = (1 - \text{exp}(\sigma_i\delta_i)),
\end{align}

\paragraph{Pruning Empty and Occluded Regions.}
However during training, finding the empty and occluded regions is not straightforward. In NerfAcc, we incorporate the idea of using an \emph{Occupancy Grid} from Instant-NGP~\cite{muller2022instant} \footnote{See Instant-NGP Appendix E.2 for details}, in which a binary grid is cached and updated during training to store which areas in the scene are empty (e.g., $\alpha_i < 1\mathrm{e}{-2}$). We skip occluded regions by terminating the marching early based on the transmittance (e.g., $T_i < 1\mathrm{e}{-4}$) along the ray, similar to Instant-NGP~\cite{muller2022instant} and PlenOctrees~\cite{yu2021plenoctrees}. Note, for skipping occluded regions, we need to know the precise density of the samples along the ray to compute transmittance. So a user-defined radiance field (e.g., MLP NeRF or Instant-NGP NeRF) is evaluated to get the density, but \emph{with gradients disabled} to minimize the computation.

\paragraph{GPU Memory Efficiency.}
One way to apply the aforementioned pruning during marching, is to first over-generate samples for each ray, and then selectively keep the samples that matter. This can be done efficiently in Python with parallelization across all the samples, but is not memory friendly as it can over allocate the GPU memory that is needed in the first place. For a scene that has dense surface such as the \emph{Lego} scene in the NeRF-Synthetic dataset, the pruning process can get rid of $98\%$ of the samples when training converges. To minimize the GPU usage, NerfAcc prunes the empty space while marching, resulting in samples only in the non-emtpy space. This changes the parallelization from samples to rays, which is slightly less efficiently but allows one to use significantly less GPU memory to generate much denser samples.

\paragraph{GPU Parallelization.}
There are in general two options for GPU parallelization: across rays or across samples. In most of the cases, the total number of valid samples in the scene is much larger than the total number of rays. As a result, we would prefer to parallelize across samples whenever possible. In NerfAcc, we minimize operations that are parallelized across rays. Operations that are independent of rays, such as computing sample opacity from sample density, are done with parallelization across samples.

\paragraph{Scene Contraction for Unbounded Scene}
As the \emph{Occupancy Grid} from Instant-NGP~\cite{muller2022instant} only supports bounded scenes\footnote{Instant-NGP proposes multi-scale grids to support a really large scenes but is still bounded.}, we incorporate the scene contraction idea\footnote{See Eq. 10 in the Mip-NeRF 360 paper} from Mip-NeRF 360~\cite{barron2022mip360} to support unbounded scenes. When querying into the occupancy grid, a non-linear function is applied to the coordinates to map the unbounded space into a finite grid. With this single change, NerfAcc is able to reconstruct bounded scenes as well as unbounded scenes.

\subsection{Differentiable Rendering}

Differentiable rendering is the process of accumulating the color of the samples along the rays, into pixel colors. In this step we make use of the user-defined radiance field in an differentiable way -- all the outputs of the radiance field will receive gradients from pixel supervisions. We follow Equation~\ref{equ:nerf} to perform the accumulation. Any sample attributes, not just colors, are supported and can be rendered in a differentiable way by NerfAcc, such as depth and opacity. We parallelize across all the samples in this step to maximize the GPU utilization.

\section{Plug-and-Play}

As the idea of NerfAcc is to perform efficient ray marching and volumetric rendering, it can work with any user-defined radiance field. To plug the NerfAcc rendering pipeline into the code and enjoy the acceleration, user only needs to define two functions (\emph{sigma\_fn}, \emph{rgb\_sigma\_fn}) with the customized radiance field. Blow is an example of plug-and-play:

\lstset{
  language=Python,
  basicstyle=\footnotesize\ttfamily,
  numberstyle=\color{gray},
  stringstyle=\color[HTML]{933797},
  commentstyle=\color[HTML]{228B22}\ttfamily,
  emph={[2]from,import,pass,return}, emphstyle={[2]\color[HTML]{DD52F0}},
  emph={[3]range}, emphstyle={[3]\color[HTML]{D17032}},
  emph={[4]for,in,def}, emphstyle={[4]\color{blue}},
  showstringspaces=false,
  breaklines=true,
  prebreak=\mbox{{\color{gray}\tiny$\searrow$}},
  numbers=left,
  frame=single,
  xleftmargin=15pt,
}
\begin{lstlisting}[language=Python]
   import torch
   from torch import Tensor
   import nerfacc 

   radiance_field = ...  # network: a NeRF model
   optimizer = ...  # network optimizer
   rays_o: Tensor = ...  # ray origins. (n_rays, 3)
   rays_d: Tensor = ...  # ray normalized directions. (n_rays, 3)

   def sigma_fn(
      t_starts: Tensor, t_ends:Tensor, ray_indices: Tensor
   ) -> Tensor:
      """ Query density values from a user-defined radiance field.
      :params t_starts: Start of the sample interval along the ray.
      :params t_ends: End of the sample interval along the ray.
      :params ray_indices: Ray indices that each sample belongs to.
      :returns The post-activation density values.
      """
      t_origins = rays_o[ray_indices]  # (n_samples, 3)
      t_dirs = rays_d[ray_indices]  # (n_samples, 3)
      positions = t_origins + t_dirs * (t_starts + t_ends) / 2.0
      sigmas = radiance_field.query_density(positions) 
      return sigmas  # (n_samples, 1)

   def rgb_sigma_fn(
      t_starts: Tensor, t_ends: Tensor, ray_indices: Tensor
   ) -> Tuple[Tensor, Tensor]:
      """ Query rgb and density values from a user-defined radiance field.
      :params t_starts: Start of the sample interval along the ray.
      :params t_ends: End of the sample interval along the ray.
      :params ray_indices: Ray indices that each sample belongs to.
      :returns The post-activation rgb and density values. 
      """
      t_origins = rays_o[ray_indices]  # (n_samples, 3)
      t_dirs = rays_d[ray_indices]  # (n_samples, 3)
      positions = t_origins + t_dirs * (t_starts + t_ends) / 2.0
      rgbs, sigmas = radiance_field(positions, condition=t_dirs)  
      return rgbs, sigmas  # (n_samples, 3), (n_samples, 1)

   # Efficient Raymarching: Skip empty and occluded space.
   packed_info, t_starts, t_ends = nerfacc.ray_marching(
      rays_o, rays_d, sigma_fn=sigma_fn, near_plane=0.2, far_plane=1.0, 
      early_stop_eps=1e-4, alpha_thre=1e-2, 
   )

   # Differentiable Volumetric Rendering.
   # colors: (n_rays, 3). opaicity: (n_rays, 1). depth: (n_rays, 1).
   color, opacity, depth = nerfacc.rendering(
      rgb_sigma_fn, packed_info, t_starts, t_ends
   )

   # Optimize the radience field.
   optimizer.zero_grad()
   loss = F.mse_loss(color, color_gt)
   loss.backward()
   optimizer.step()
\end{lstlisting}

\section{Example Usages}

We provide four example usages of NerfAcc on different radiance field representations, on different type of scenes, including dynamic scenes and unbounded scenes. The results below indicate that NerfAcc is a universal acceleration tool for various NeRFs. Our experiments are all conducted on a single NVIDIA TITAN RTX GPU.

\subsection{Vanilla MLP NeRF in 1 hour.}
In this example we trained an 8-layer-MLP with the same structure as the one in the original NeRF~\cite{mildenhall2021nerf} method. We used the train split for training and test split for evaluation as in the NeRF paper. The training memory footprint is about 10GB. The results on the NeRF-Synthetic dataset is shown here:
\input{tables/nerf}
\paragraph{Note.} The vanilla NeRF paper uses two MLPs for course-to-fine sampling. Instead here we only use a single MLP with more samples (1024). Both methods accomplish the same goal of dense sampling around the surface. NerfAcc inherently skips samples away from the surface so we can simply increase the number of samples with a single MLP, to achieve the same goal with the coarse-to-fine sampling, without run-time or memory issue.

\subsection{Instant-NGP in 4.5 minutes.}
In this example we trained a Instant-NGP NeRF model on the NeRF-Synthetic dataset. We adopt the same settings proposed in the Instant-NGP paper, which uses train split for training and test split for evaluation. The training memory footprint is about 3GB.
\input{tables/ngp}
\paragraph{Note.} The Instant-NGP paper makes use of the alpha channel in the images to apply random background
augmentation during training. For fair comparison, we rerun their code with a constant white
background during both training and testing. Also it is worth to mention that we didn't strictly
follow the training recipe in the Instant-NGP paper, such as the learning rate schedule etc, as
the purpose of this benchmark is to showcase instead of reproducing the paper.

\subsection{D-NeRF in 1 hour.}
In this example we trained a 8-layer-MLP for the radiance field and a 4-layer-MLP for the warping field, (similar to the T-NeRF model in the D-NeRF paper) on the D-NeRF dataset. We used train split for training and test split for evaluation. The training memory footprint is about 11GB.
\input{tables/dnerf}
\paragraph{Note.} The Occupancy Grid used in this example is shared by all the frames. In other words, instead of using it to store the opacity of an area at a single timestamp, Here we use it to store the maximum opacity at this area over all the timestamps. It is not optimal but still makes the rendering very efficient.

\subsection{Unbounded Scene in 20 minutes.}
Here we trained an Instant-NGP NeRF on the Mip-NeRF 360 dataset. We used the train split for training and test split for evaluation. The training memory footprint is about 8GB.
\input{tables/unbounded}
\paragraph{Note.} Even though we are comparing with NeRF++ and Mip-NeRF 360, the model and losses are different. There are many ideas from those papers that could be helpful for the performance. We didn’t adopt them in these experiments as they are tangential to the library.

\section{Conclusion}
We have demonstrated a general acceleration toolbox, NerfAcc, for accelerating various NeRFs in different applications. It is built on top of the ideas from Instant-NGP~\cite{muller2022instant}, Plenoctrees~\cite{yu2021plenoctrees} and Mip-NeRF 360~\cite{barron2022mip360} for efficient volumetric rendering. With NerfAcc, various NeRF models can be trained in 1 hour with better performance. We provide full Python API for NerfAcc, which makes it ready for plug-and-play in future research.


\bibliographystyle{unsrt}  
\bibliography{references}

\end{document}

%% file: tables/nerf.tex
\begin{table}[h]
\begin{center}
\resizebox{1.0\linewidth}{!}{
\begin{tabular}{l c c c c c c c c c c}
\hline\hline\noalign{\smallskip}
& Lego & Mic & Materials & Chair & Hotdog & Ficus & Drums & Ship & Mean \\
\hline
\noalign{\smallskip}
NeRF~\cite{mildenhall2021nerf} (PSNR: $\sim$days) & 32.54 & 32.91 & 29.62 & 33.00 & 36.18 & 30.13 & 25.01 & 28.65 & 31.00 \\
Ours (PSNR: $\sim$1 hr) & 33.69 & 33.76 & 29.73 & 33.32 & 35.80 & 32.52 & 25.39 & 28.18 & 31.55 \\
Ours (Training time) & 58min & 53min & 46min & 62min & 56min & 42min & 52min & 49min & 52min \\
\hline\hline
\end{tabular}
}
\end{center}
\caption{\textit{Performance on NeRF-Synthetic dataset}.}
\end{table}

%% file: tables/ngp.tex
\begin{table}[h]
\begin{center}
\resizebox{1.0\linewidth}{!}{
\begin{tabular}{l c c c c c c c c c c}
\hline\hline\noalign{\smallskip}
PSNR & Lego & Mic & Materials & Chair & Hotdog & Ficus & Drums & Ship & Mean \\
\hline
\noalign{\smallskip}
Instant-NGP~\cite{muller2022instant} ($\sim$ 4.5 mins) & 35.87 & 36.22 & 29.08 & 35.10 & 37.48 & 30.61 & 23.85 & 30.62 & 32.35 \\
Ours ($\sim$ 4.5 mins) & 35.50 & 36.16 & 29.14 & 35.23 & 37.15 & 31.71 & 24.88 & 29.91 & 32.46\\
\hline\hline
\end{tabular}
}
\end{center}
\caption{\textit{Performance on NeRF-Synthetic dataset}.}
\end{table}\textbf{}

%% file: tables/dnerf.tex
\begin{table}[h]
\begin{center}
\resizebox{1.0\linewidth}{!}{
\begin{tabular}{l c c c c c c c c c}
\hline\hline\noalign{\smallskip}
PSNR & Bouncing balls & Hell warrior & Hook & Jumping jacks & Lego & Mutant & Standup & Trex & Mean \\
\hline
\noalign{\smallskip}
D-NeRF~\cite{pumarola2021d} ($\sim$ days) & 38.93 & 25.02 & 29.25 & 32.80 & 21.64 & 31.29 & 32.79 & 31.75 & 30.43 \\
Ours ($\sim$ 1 hr) & 39.49 & 25.58 & 31.86 & 32.73 & 24.32 & 35.55 & 35.90 & 32.33 & 32.22 \\
Ours (Training time) & 37min & 52min & 69min & 64min & 44min & 79min & 79min & 39min & 58min \\
\hline\hline
\end{tabular}
}
\end{center}
\caption{\textit{Performance on D-NeRF Synthetic dataset}.}
\end{table}\textbf{}

%% file: tables/unbounded.tex
\begin{table}[h]
\begin{center}
\resizebox{1.0\linewidth}{!}{
\begin{tabular}{l c c c c c c c c}
\hline\hline\noalign{\smallskip}
PSNR & Garden & Bicycle & Bonsai & Counter & Kitchen & Room & Stump & Mean \\
\hline
\noalign{\smallskip}
NeRF++~\cite{zhang2020nerf++} ($\sim$ days) & 24.32 & 22.64 & 29.15 & 26.38 & 27.80 & 28.87 & 24.34 & 26.21 \\
Mip-NeRF 360~\cite{barron2022mip360} ($\sim$ days) & 26.98 & 24.37 & 33.46 & 29.55 & 32.23 & 31.63 & 28.65 & 29.55 \\
Ours ($\sim$ 20 mins) & 25.41 & 22.97 & 30.71 & 27.34 & 30.32 & 31.00 & 23.43 & 27.31\\
Ours (Training time) & 25min & 17min & 19min & 23min & 28min & 20min & 17min & 21min\\
\hline\hline
\end{tabular}
}
\end{center}
\caption{\textit{Performance on Mip-NeRF 360 dataset}.}
\end{table}\textbf{}